\documentclass[nohyperref]{article}

\usepackage{microtype}
\usepackage{graphicx}
\usepackage{booktabs} % for professional tables

\usepackage{amsmath}
\usepackage{amsthm}
\usepackage{amsfonts}
\usepackage{dsfont}
\usepackage{hyperref}

\usepackage{color}

\usepackage{caption}
\usepackage{graphicx}

\usepackage{caption}
\usepackage{subcaption}

\usepackage[table]{xcolor}% http://ctan.org/pkg/xcolor

\usepackage[decisionutilitycolor]{influence-diagrams}

\usepackage[accepted]{icml2022}

\usepackage{amsmath}
\usepackage{amssymb}
\usepackage{mathtools}
\usepackage{amsthm}

\usepackage[capitalize,noabbrev]{cleveref}

\theoremstyle{plain}
\newtheorem{theorem}{Theorem}[section]

\theoremstyle{definition}
\newtheorem{definition}[theorem]{Definition}

\theoremstyle{remark}

\usepackage[textsize=tiny]{todonotes}

\icmltitlerunning{Argumentative Reward Learning}

\begin{document}

\twocolumn[
\icmltitle{Argumentative Reward Learning: Reasoning About Human Preferences}

% submissions: the style file  automatically remove it for you

% identifier you  use later to specify author affiliations

% Ideally, you should not use this facility. Affiliations  be numbered
\icmlsetsymbol{equal}{*}

\begin{icmlauthorlist}
\icmlauthor{Francis Rhys Ward}{y}
\icmlauthor{Francesco Belardinelli}{y}
\icmlauthor{Francesca Toni}{y}
\end{icmlauthorlist}

\icmlaffiliation{y}{Department of Computing, Imperial College London}

\icmlcorrespondingauthor{Francis Rhys Ward}{francis.ward19@ic.ac.uk}

% the "keywords" metadata in the PDF but  not be shown in the document
\icmlkeywords{Machine Learning, ICML}

\vskip 0.3in
]



\printAffiliationsAndNotice{} % otherwise use the standard text.
% 4 pages in length (excluding references and acknowledgments)
% 16th May
% Abstracts must be a single paragraph, ideally between 4--6 sentences long.
% Gross violations  trigger corrections at the camera-ready phase.

% \nb{include reward heatmaps}

\begin{abstract}

% In settings without well-defined goals, methods for reward learning allow reinforcement learning agents to infer the goal from human feedback. In an online setting,  the reward model can be updated via further interaction with the user, to correct misunderstandings, or if the objective changes. Non-monotonic logics are therefore well suited to reasoning about human preferences online, allowing the agent to draw conclusions defeasibly with the ability to retract these conclusions under the light of further interaction with the human. We use PBAto represent and reason  non-monotonically about human preferences. 
We define a novel neuro-symbolic framework, \emph{argumentative reward learning,} which combines preference-based argumentation with existing approaches to reinforcement learning from human feedback. %In this framework, trajectories in an MDP are represented as arguments and (preference-based) argumentative semantics are then used to reason about human preferences over agent behaviour. 
Our method improves prior work by generalising human preferences, reducing the burden on the user and increasing the robustness of the reward model. We demonstrate this with a number of experiments. 
\end{abstract}

\section{Introduction}
\label{sec:intro}

Recent success has been made in (deep) reinforcement learning (RL) in settings with well-defined goals (e.g., achieving expert human level in Atari games ~\cite{mnih2013playing}, Go ~\cite{silver2016mastering}, Starcraft ~\cite{VinyalsBCMDCCPE19}). However, RL has had limited success with real-life tasks for which the goals are not easily specified ~\cite{Whittlestone}, leading to a body of work on %the {\em AI alignment problem}: the problem of aligning the goals (as expressed by the reward function) with the intent of the designers or users. Hence, methods of 
\emph{RL from human feedback (RLHF)} %have been proposed as a solution to the alignment problem in which the reward function is also taken as something to be learned ~\cite{leike_scalable_2018,hadfieldmenell2016cooperative,christiano_deep_2017}. 
in which the reward function is also learned from data about human preferences ~\cite{NIPS2017_7017,leike_scalable_2018,hadfieldmenell2016cooperative}. 

Past work on RLHF typically utilizes supervised deep learning (DL) to predict human preferences~\cite{NIPS2017_7017,ziegler2019fine}. We address two challenges for a pure DL approach in this setting. Firstly, no explicit reasoning occurs about the cause of a given preference, meaning that the model can learn spurious correlations or otherwise fail to generalize from the training set~\cite{marcus2018deep}. Secondly, DL is data-inefficient, meaning that DL-based approaches are bottle-necked by data about human preferences, which is costly to collect.
%Abstract \mathcal{A}umentation ~\cite{dung} is a non-monotonic logic and therefore well-suited for reasoning defeasibly in an online learning setting. 
In contrast to DL, preference-based argumentation  (PBA)~\cite{modgil2009reasoning,amgoud2013acceptability} offers a natural method for reasoning about human preferences. 
%In general, PBAallows to resolve conflicts between available information with the help of preferences over alternatives. 
It supports non-monotonic reasoning,  
%Non-monotonic logics are devised to capture and represent defeasible inferences, i.e., a kind of inference 
whereby reasoners draw tentative conclusions, %enabling them to retract their conclusion 
which may be retracted based on further evidence. 
This form of reasoning is well-suited to online RLHF because 
%In an online RLHF setting,  
the reward model can be updated via further interaction with the user, to correct misunderstandings, or if the objective changes. %Non-monotonic logics are therefore well suited to reasoning about human preferences online, allowing the agent to draw conclusions defeasibly with the ability to retract these conclusions under the light of further interaction with the human. 
We use PBA to represent and reason  non-monotonically about human preferences,
allowing the agent to draw conclusions defeasibly, with the ability to retract these conclusions under the light of further interaction with the human.

\emph{Contribution.} We present a novel  neuro-symbolic \cite{vismeta} 
method  which
integrates %argumentative reasoning 
PBA into RLHF.
Specifically, we see trajectories as arguments disagreeing with dissimilar trajectories and inject human preferences over trajectories to enrich RL. 
We show how 
this ameliorates the challenges %for 
faced by a pure DL approach. % by using argumentative semantics to reason about and generalise a small amount of HF
The rest of this paper is structured as follows. In the next section, we present %the
background on \emph{RL}, \emph{RLHF}, and \emph{PBA}. In section \ref{sec:theory_steps} we present our framework for \emph{argumentative reward learning (ARL)}. In sections \ref{sec:cont} and \ref{sec:results} we describe our experiments and results in a maze-solving task. We conclude with a discussion. 

\section{Background}

\emph{Reinforcement learning (RL)} \cite{sutton2018reinforcement} is well-known%, we define the
. We use the notion of  %(fully-observable, finite, deterministic) 
Markov decision process% (MDP)
.

\begin{definition}%[MDP]
\label{def:mtag}
A \emph{Markov decision process (MDP)} is a tuple $%M=
(S,A,T(),P(), r(), \gamma), \text{ where: }$ $S$ is a set of \emph{states} and $A$ is a set of \emph{actions}; 
    %  $T: S \times A \times S \mapsto [0,1] $ gives a \emph{conditional probability} $T(s' \mid s,a)$ on the next world state $s'$, given the previous state $s$ and action $a$;
    $T: S \times A  \to S $ is a (deterministic) function mapping a state $s$ and action $a$ to the next world state $s'$;
     $P: S  \to [0,1]$ is a \emph{prior probability distribution} over the initial state%\ft{ IS THERE A SINGLE INITIAL STATE?}
     ;
     $r:S\times A \times S  \to \mathbb{R}$ 
 is a reward function that maps a transition $(s, a, s')$ to a real number;
  $\gamma\in [0,1]$ is the \emph{discount factor}.

% \begin{itemize}
%     \item $S$ is a set of \emph{states} and $A$ is a set of actions; 
%     \item $T: S \times A \times S \mapsto [0,1] $ gives a \emph{conditional probability} $T(s' \mid s,a)$ on the next world state $s'$, given the previous state and action;
%     \item $P: S  \mapsto [0,1]$ is a prior probability distribution over the initial state;
%     \item $R:S\times A \times S  \mapsto \mathbb{R}$ 
%  is a reward function that maps a transition $(s, a, s')$ to a real number;
%  \item $\gamma\in [0,1]$ is the \emph{discount factor}.
% \end{itemize}

\end{definition}

In an MDP, a \emph{trajectory} is a finite sequence of state-action pairs:  $\tau = (s_0, a_0), (s_1, a_1), \ldots, (s_N, a_N)$ (where $s_0$ may be any state)%. We also define 
; the \emph{return}, $R$, of a trajectory is the total (time-discounted) reward earned:  $R_{\gamma}(\tau) = \sum_{i=0}^{i=N} \gamma^i r(s,a, s') $%, with discount rate $\gamma$ ALREADY IN DEF 1
; for $\gamma =1$, we drop the subscript and write $R(\tau)$.
%Behaviour in the game 
Finally, agent behaviour is described by a \emph{policy} $\pi:S  \to A$. % A policy is \emph{optimal} if it maximises the expected return.

\emph{RLHF} \label{sec:rlhf}
%There 
is the focus of a rich literature %on RLHF 
\cite{NIPS2017_7017,NIPS2012_4805,akrour2011preference,frye2019parenting,ak,wang2016learning,rahtz2019extensible} %and demonstrations \cite{hester2017deep, bain1995framework, ho2016generative, schaal1999imitation}. 
%In recent years, RLHF 
and has become, in recent years, a standard training regime, particularly for large language models ~\cite{ziegler2019fine,stiennon2020learning,ouyang2022training,wu2021recursively,rae2021scaling}. 
% has aimed to communicate complex goals to RL systems by expressing human preferences over pairs of example trajectories through the environment, thus enabling the system to infer an estimate of an underlying reward function from sub-optimal agent behaviour. We can thus train a reward-model using supervised learning to predict these preferences. This can be used in scenarios where producing expert demonstrations is difficult. This approach has four steps:
In section \ref{sec:theory_steps} we extend the method of \cite{NIPS2017_7017} to incorporate argumentative reasoning. %which is as follows: 1) Initialise a policy $\pi$ and use this to sample trajectories from the environment; 2) Sample pairs of trajectories and query a human-in-the-loop to state a preference between them; 3) Optimize a reward model $r$ to fit the human preferences using supervised-learning; 4) Update $\pi$ using $r$ with a standard RL algorithm. In section ~\ref{sec:theory_steps} we discuss this process, with our added contributions, in detail. 

% However, the third step has so far learned to predict human preferences in a black-box way, using neural networks (NNs) - no \textit{reasoning} about human preferences occurs. %And the preference relation is treated as a binary prediction problem, with no representation of how much better one behaviour is than another and no explanation of why.
% Another serious issue with this process is it's scalability - eliciting information from human preferences in this way is inefficient and requires large amounts of human time. Using abstract argumentation we  ameliorate this issue by generalizing human preferences provided by direct feedback. %(using a method of preference derivation similar to \cite{10.1007/978-3-030-29908-8_55}). 

% Further work extends the process by including policy pre-training stages which train the policy via imitation learning on expert demonstrations \cite{ibarz2018reward}. They also integrate these demonstrations with the preference learning by automatically pairing demonstrations with a random trajectory and labelling the demonstration as preferred. In section \ref{sec:exp}, we  similarly utilise demonstrations by taking them as ``acceptable" arguments, from which we can derive preferences in a more principled way. 

% There is limited other research that combines learning from both preferences and demonstrations in this way.

\label{sec:Arg}
\emph{Argumentation} is a well-known area of symbolic AI (see \cite{Atkinson2017Oct} for a recent overview). We use a form of argumentation accommodating preferences into the popular abstract argumentation paradigm \cite{dung}.

\begin{definition}
An \emph{abstract argumentation framework} (AAF) \cite{dung}  is a pair $(\mathcal{A},\mathcal{R})$ where $\mathcal{A}$ is a given set (whose elements represent \emph{arguments}) and $\mathcal{R} \subseteq \mathcal{A} \times \mathcal{A}$ (referred to as the \emph{attack relation}). A set of arguments $S \subseteq \mathcal{A}$ \emph{attacks} $A \in \mathcal{A}$ iff $(B, A) \in \mathcal{R}$ for some $B \in S$.
\end{definition}

Argumentation frameworks are equipped with a
\emph{semantics} determining which arguments are dialectically good% \cite{}
. We focus on the \emph{preferred semantics} for AAFs~\cite{dung}.

\begin{definition}%[Argumentative Semantics] 
For an AAF $(\mathcal{A}, \mathcal{R})$, let
 $A \in \mathcal{A}$ be \emph{acceptable} w.r.t. $S \subseteq \mathcal{A}$ if and only if $S$ attacks  every $B \in \mathcal{A}$ such that $(B,A) \in \mathcal{R}$, and let % attacking $A$ 
 $S\subseteq \mathcal{A}$ be \emph{conflict-free} if and only if there exist no
 $A,B \in S$ such that
 $(B,A) \in \mathcal{R}$% no two arguments in $S$ attack each other
 . Then, 
for any conflict free $S \subseteq \mathcal{A}$, $S$ is an \emph{extension} that is:
     \emph{admissible} if every argument in $S$ is acceptable w.r.t. $S$;
     \emph{preferred} if it is maximally (under set inclusion) admissible. % and every argument acceptable w.r.t. $S$ is in $S$.
\end{definition}
%\emph{Preferred extensions} can be seen as sets of arguments representing a desirable position in the debate.

PBA offers a natural way to reason about human preferences \cite{modgil2009reasoning,zeng2020explainable}. %We can reduce a PAF to an AAF using the \emph{PAF reduction}.

\begin{definition}
A \emph{Preference-based AAF (PAF)} \cite{amgoud2013acceptability} is a triple $(\mathcal{A},\mathcal{R}, \succ)$ where $(\mathcal{A}, \mathcal{R})$ is an AAF and $\succ \mathcal{A} \times \mathcal{A}$ is a transitive and asymmetric binary relation on $\mathcal{A}$; $A \succ B$ means that $A$ is preferred to $B$.
\end{definition}

 PAFs $(\mathcal{A},\mathcal{R}, \succ)$ can be reduced to AAFs %using the \emph{PAF reduction}. The reduction of a PAF is $AAF_{\succ} = 
 $(\mathcal{A},\mathcal{R}_{\succ})$ where $\mathcal{R}_{\succ}= \mathcal{R} \setminus \{(B, A)|A \succ B\}$ (that is, %we reduce the argumentation framework 
 by removing attacks from $B$ to $A$ for all $A \succ B$).

\section{Method: Argumentative Reward Learning} \label{sec:theory_steps}

\begin{figure}[t!]
\centering
\includegraphics[width=0.5\textwidth]{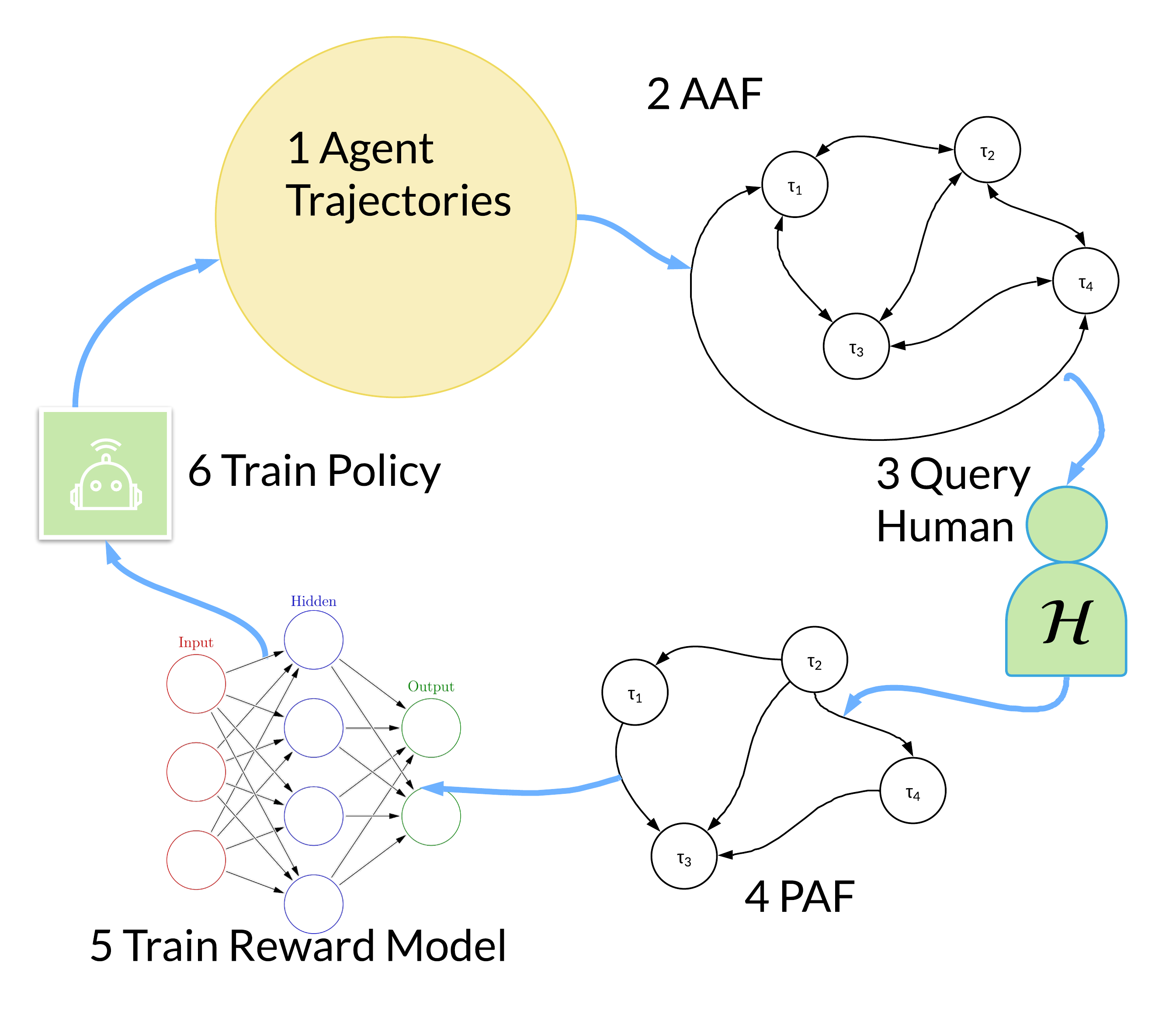}
\vspace{-10mm}
\caption{Argumentative reward learning (ARL) framework. Step 1: Collect agent trajectories. Step 2: Define the AAF (visualised here as a graph, with arguments/trajectories as nodes and attacks as edges). Step 3: Query human preferences. Step 4: %Generalise the preferences.
Obtain the PAF. Step 5: Train the reward model. Step 6: Train the policy.}
\label{fig:process}
\end{figure}

Here we present \emph{ARL}, which %takes 
interprets trajectories in an MDP as arguments in an AAF and uses %argumentative
preferred
 semantics to %reason about preferences in order to 
 generalise HF.
We start with an agent interacting with an MDP environment to generate trajectories. We then define an AAF from these trajectories and generate a PAF by using and generalising the HF. We train a reward model on %the attacks of 
this PAF and then use this reward model to train the RL agent.
 Hence, we split the method into six steps as seen in Fig. \ref{fig:process}.

% \begin{enumerate}
%     \item The agent generates trajectories in the environment using some policy, $\pi$. %Collect expert demonstrations of acceptable/non-acceptable trajectories.
%     \item Define an argumentation framework, $(\mathcal{A}, \mathcal{R})$, with trajectories as arguments and an attack relation based on similarity between trajectories.
%     \item Query the human, $\mathcal{H}$, with either a pair of attacking trajectories or a single trajectory. Given a pair, $\mathcal{H}$ may label a preference, given a single trajectory $\mathcal{H}$ may label ``acceptable"/ ``not acceptable"
%     \item Generate the PAF $(\mathcal{A}, \mathcal{R}_{\succ})$ from $(\mathcal{A}, \mathcal{R})$ using the human preferences and semantics/generalisation
%     \item Optimize the reward model, $\hat{r}$ via supervised learning to fit the attacks $\mathcal{R}_{\succ}$ (attacks in $\mathcal{R}_{\succ}$ correspond to either labelled or inferred preferences)
%     \item Train $\pi$ to maximises $\hat{r}$ using RL
% \end{enumerate}

% Our contribution is steps 2 and 4; step 2 ensures that the queries to the human are more useful, and step 4 gives the main contribution of generalising the feedback. We now go through each step in detail. 

\emph{Step 1: Collecting Agent Trajectories.} We start with a randomly initialised policy~\cite{NIPS2017_7017}.% (an alternative is to start with expert demonstrations as in ~\cite{ibarz2018reward}). %(an extension of this work could start with expert demonstrations as in \cite{ibarz2018reward}).

\emph{Step 2: Defining the %Argumentation Framework
AAF.} \label{sec:daf} We now take \emph{trajectories} to be \emph{arguments} in an AAF. We define the \emph{attack} relation based on a notion of similarity between trajectories: %. In this way, 
dissimilar trajectories %which
are deemed to ``disagree" about the correct behaviour %are 
and are thus  taken to be mutually attacking arguments.

\emph{Step 3: Querying %Preferences
the Human.} We elicit preferences over trajectories from the human in the loop, by querying 
% Once we have our argumentation framework, and a set of attacks between trajectories, we wish to derive the preference-based framework; hence we need to query the human-in-the-loop for their preferences over trajectories. 
%Past work samples trajectories either uniformly at random \cite{ibarz2018reward} or based on the uncertainty in the reward model \cite{NIPS2017_7017}. In contrast, we query
pairs of \emph{attacking} trajectories, ensuring that queried trajectories actually exhibit different behaviour (differently from past work on preference elicitation, which samples trajectories uniformly at random \cite{ibarz2018reward} or based on the uncertainty in the reward model \cite{NIPS2017_7017}).

% Indeed, these previous approaches include an option to label the queried trajectories as ``incomparable" whereas this should not be necessary in our framework (although we may include an option for ``no preference"). 

% Other work \cite{ibarz2018reward} combining learning from preferences and demonstration sometimes automatically labels the demonstrations as more preferred to all agent trajectories. We may also include the demonstrations in our AAF and take them as more preferred to the arguments which they attack. Otherwise we may label them as ``acceptable" with respect to some AA semantics (e.g. so that they represent one preferred extension and one policy in the environment). Similarly we may include an option for the human to label trajectories as acceptable/non-acceptable. These labels can then be used in a preference derivation similar to that of \cite{10.1007/978-3-030-29908-8_55} to generalise the preferences to other trajectories.

\emph{Step 4: %Generalising the Preferences
Defining the PAF.}
% A key idea is that, after observing some human preferences, AI should reason about these and extrapolate. In \cite{10.1007/978-3-030-29908-8_55} a process is outlined to infer preferences over a set of arguments from a smaller subset of labelled preferences (where we can label both arguments and attacks as accepted). As shown in section \ref{sec:prelim}, we can use a similar process to generalise human preferences over trajectories.
The human preferences collected at Step 3 can be used as preferences over arguments. %(assuming that transitivity and asymmetry are natural properties satisfied by humans). 
However, with limited HF these preferences are only partial. In order to generalise them, 
we
use the preferred %extension
semantics of the AAF to determine preferred sets of trajectories.  % generalise the stated human preferences to a larger set of trajectories. 
Because of our notion of attack, these preferred extensions %correspond to sets of
only include similar trajectories. 
Then, we determine a preference order $\gg$ over preferred extensions (see Section~\ref{sec:results} for details). Finally, for
trajectories $\tau_1, \tau_2$ in the set of arguments in the AAF,
 we take $\tau_1 \succ \tau_2$ iff $\exists p_1$ (a preferred extension of the AAF) with $\tau_1 \in p_1$ such that $\forall p_2$ (also preferred extensions of the AAF) with $\tau_2 \in p_2$ it holds that $p_1\gg p_2$.
% This way,  we generalise the HF to a larger set of trajectories. 
%then we use the ordered extensions to get preferences over all args, 
This results in a PAF, which can be reduced to  $(\mathcal{A},\mathcal{R}_{\succ})$ (as defined in Section~\ref{sec:Arg}). 
% \nb{could there be inconsistencies: that a pref extension is both more and less preferred than another? what do you do in this case? Can't happen!}
% Note that, t
This process is a simple way to generalise the stated human preferences (i.e. a small amount of HF) to a larger set of (similar) trajectories: more sophisticated ways to generalise the preferences are left for future work.
%
%\nb{I have moved below something you had later in the experiments. It does not address the issue with preferred extensions - and worries me that actually you are not using at all the semantics - thus not generalising the HF?}
%\ft{ We take each attack in the PAF as a preference over trajectories, i.e., for all $(\tau_1,\tau_2) \in \mathcal{R}_{\succ}$ we take $\tau_1$ as \textit{preferred} to $\tau_2$.\nb{preferred is a very overloaded term here...} 
Next, we train the reward model to predict these preferences.

%and we can therefore use a small amount of HF to elicit an ordering over preferred extensions which  generalise this feedback.

% Once we have generalised the human preferences, we are in a position to train a reward model.

\emph{Step 5: Training the Reward Model%$\hat{r}$
.} \label{sec:5}
As in \cite{NIPS2017_7017}, we  train a neural network to predict the immediate reward,  $\hat{r}: S \times A \rightarrow \mathbb{R}$, %given a state and an action,
based on the preferences derived in
the previous step (we use $\hat{r}$ to distinguish a learned reward model  from the \emph{true reward function} $r$.  We assume the preferences between trajectories drawn from Step 3 %is a function 
are functions of the reward that we aim to obtain at this step, such that the probability %that 
of $\tau_1 \succ \tau_2$ is given by     

$
    \text{Pr}(\tau_1 \succ \tau_2) = \frac{\exp(\hat{R}(\tau_1))}{\exp(\hat{R}(\tau_1)) + \exp(\hat{R}(\tau_2))},
$ 
%  $
%     P(\tau_1 \succ \tau_2) = \frac{\exp(R(\tau_1))}{\exp(R(\tau_1)) + \exp(R(\tau_2))}
%   $
where $\hat{R}(\tau)$%=\ft{\hat{R}%_1}(\tau)$ 
is the non-discounted return, using $\hat{r}$, for trajectory $\tau$ (i.e. $\hat{R}_\gamma$ for $\gamma = 1$). We then 
%optimize 
%our reward model, 
obtain $\hat{r}%: S \times A \rightarrow \mathbb{R}
$ %to minimize 
by minimizing the binary cross-entropy loss between these predictions and the %attacks 
preferences between trajectories in the PAF:    
$
    \text{loss}(\hat{r}) = - \sum_{(\tau_1, \tau_2) \in \mathcal{R}_{\succ}} \log  \text{Pr}(\tau_1 \succ \tau_2).
$ 
% $ \label{eq:ploss}
%   \text{loss}(\hat{r}) = - \sum_{(\tau_1, \tau_2) \in \mathcal{R}_{\succ}} \log  P(\tau_1 \succ \tau_2)
% $
%where $\tau_1 \succ \tau_2$ \todo{what does this mean? $\tau_1 \succ \tau_2$ belong somewhere? also, what is the intutition that we consider attacks between trajectories here?}.
%This follows the BT model \cite{10.1093/biomet/39.3-4.324}  for estimating score functions from pairwise preferences. %(analogous to ELO rankings in chess ~\cite{elo1978rating}).  %It can be interpreted as equating rewards with a preference ranking scale, analogous to the Elo ranking system developed for chess \cite{elo1978rating}.

% \todo[inline]{in the experimental part you seem to indicate that we use preferences over preferred extensions - nothing is mentioned here...I am very confused...}

\emph{Step 6: Training the Policy%$\pi$
.} Once we have trained $\hat{r}$ we are left with a standard RL problem. To train the policy $\pi$ we use deep Q-learning \cite{mnih2015human}.%, the full algorithm and experimental details are included in appendix section \ref{sec:}.

% \subsection{Summary}

% In this section we have outlined the high-level method used in this report, and we also discussed how this method is one particular approach to a more general solution to the preference learning problem (i.e. the general process of defining a formalism to reason about human preferences, learning this formalism, and then using it to train behaviour). Our method builds on previous work by including Steps 2 and 4, that is by defining an attack relation between trajectories (thus making the queries more useful) and by generalising the HF using the argumentative semantics. In the next chapter we perform some experiments to determine whether our improvements lead to any performance differences compared to past work (e.g. in terms of the number of queries needed) and to standard Deep RL (in terms of the resulting policy). We first look at a discrete setting and then extend our approach to a continuous maze environment. 

\section{Experiments: Maze Solving } \label{sec:cont}

\begin{figure}[t!] 
\centering
\includegraphics[width=0.25\textwidth]{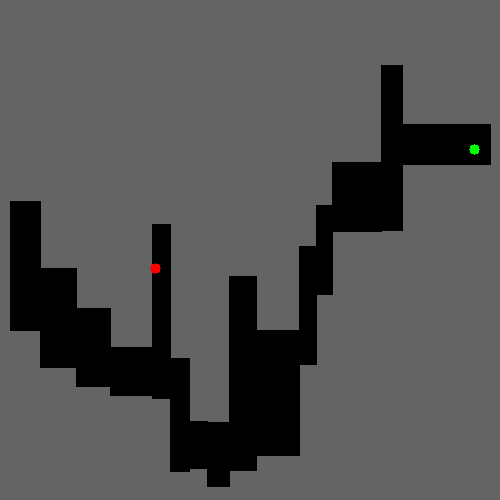}
\caption{Maze environment: red agent searches for the green goal, navigating unpassable walls. %All the experiments presented are conducted on the map shown here.
}
\label{fig:ded}
\end{figure}

\emph{Environment definition. } We conduct a number of experiments in a maze-solving MDP environment (Fig. \ref{fig:ded}), defined by:  
     %the \emph{state space:} 
     $S = \{(x,y) | x, y \in [0,1] \}$, including randomly generated walls;
     %the \emph{action space:} 
     $A = \{(0,0.02),(0.02,0),(0,-0.02),(-0.02,0)\}$ (i.e. up, right, down, left); and $T()$ and $r()$ given by
      %   the \emph{transition function:} 
    %   $T: S \times A \rightarrow S$
      \newline $
          T(s,a) %\mapsto
      =
        \begin{cases}
        s, \text{ if $s'$ is a wall} \\
        s', \text{ otherwise}
        \end{cases}  \text{where}  s' := s+a;
$     %the \emph{reward function:} 
    % and $r: S \times A \times S \rightarrow \mathbb{R}$ given by
    
      $
    r(s,a,s') %\mapsto
    =
    \begin{cases}
      1,  \text{ if } d \leq 0.3, \\
      (1- d)^2 -0.1,  \text{ if } s = s', \\
      (1- d)^2,  \text{ otherwise}
      
    \end{cases}$
  
where $d$ is the Euclidean distance from $s'$ to the goal, normalised to [0, 1]. Note that this reward function is the result of approximately $10$ hours of reward design in combination with other hyper-parameter tuning.  %(e.g. of the RL parameters: trajectory length, $\epsilon$ level of exploration etc, discussed in detail in section \ref{sec:hyper})in order to develop an effective RL procedure for maze-solving in these randomly generated maze environments. 

% \begin{itemize}
%     \item The \emph{state space:} $S = \{(x,y) | x, y \in [0,1] \}$, including randomly generated, inaccessible walls,
%     \item The \emph{action space:} $\{(0,0.02),(0.02,0),(0,-0.02),(-0.02,0)\}$ (i.e. up, right, down, left),
%         \item The \emph{transition function:} $T: S \times A \rightarrow S$, \\ $T(s,a) \mapsto 
%         \begin{cases}
%         s, \text{ if $s'$ is a wall} \\
%         s', \text{ otherwise}
%         \end{cases}$, where $s' := s+a$,
%     \item The \emph{reward function:} $r: S \times A \times S \rightarrow \mathbb{R}$, 
%     $
%     r(s,a,s') \mapsto
%     \begin{cases}
%       1,  \text{ if } d \leq 0.3, \\
%       (1- d)^2 -0.1,  \text{ if } s = s', \\
%       (1- d)^2,  \text{ otherwise.}
      
%     \end{cases}
%   $
% \end{itemize}

\emph{Experiment outline.} We train %a number of 
eight reward models for the continuous maze environment in Fig.~\ref{fig:ded}. We apply  the method of \citet{NIPS2017_7017} to train four benchmark models: one from synthetic preferences, %one
two from %both
100 and 200 human preferences, respectively, and one trained iteratively by cycling through the steps 1-6.  We train four analogous models using our ARL method. 

\emph{Step 1%: Collecting Agent Experience
.} We first generate $100$ random trajectories in the maze; these trajectories start from random initial states and have length $20$. %We start from random initial states as this efficiently explores the state space which  allow us to train the reward model more easily. 
% Here we train the reward model only once rather than iteratively as in later experiments.

\emph{Step 2%: Defining the Argumentation Framework
.}  From these initial trajectories we generate an AA framework with the (symmetric) attack relation:     %$\mathcal{R}(\tau_1, \tau_2) 
$(\tau_1, \tau_2)  \in \mathcal{R}$  iff  $\exists i \in \{0,..., N\}$  s.t.  $||s_1(i) - s_2(i)|| > \delta$,
% \begin{align}
%     \tau_1 \text{ attacks } \tau_2 \text{ iff } ||s_1(i) - s_2(i)|| > \delta \text{ for } i \in [0,T].
% \end{align}
where $s_1$ and $s_2$ are the sequences of states in $\tau_1$ and $\tau_2$ respectively, $s_j(i)$ is the $i$-th element of sequence $s_j$ (for $j=1,2$), $||.||$ is the Euclidean distance, $N=19$,
and the threshold value of $\delta$ is heuristically chosen (for this experiment we use $\delta = 0.2$). 
% \todo[inline]{above very underspecified: what are $s_1$ and $s_2$? how is $i$ quantified (for all? there exists)? what is $T$ (I guess not the transition function, and I guess it is actually 19)?
% should $[0,T]$ be $\{0, \ldots, T\}$ (discrete, rather than continuous, set)? how are any of these symbols connected to $\tau_1$ and $\tau_2$? Also, note that you use $d$ for Euclidian distance before - but to the goal state and normalised - so ok I guess to use a different notation here }
% and a trajectory length of $T = 20$.
This attack relation captures a form of similarity between trajectories, %i.e. 
whereby if %they
two trajectories are close to each other at every time step then they do not attack one another. From our $100$ trajectories, with our choice of parameters, we get an AAF with $8230$ attacks.% and $105$ preferred extensions. 
% \nb{does it matter whether we have a single environment as in fig. 2 or different ones? do we always get the same attacks? I guess some trajectories will not be allowed in some environments - due to walls, so are you just considering a single env as in fig 2? a comment earlier on would be beneficial}
% Some of these extensions are shown in figure \ref{fig:cont_ps}. 

\emph{Step 3%: Querying Preferences
.}
We consider two settings, respectively with synthetic preferences and preferences drawn from HF.

\emph{Synthetic Preferences. }As in \cite{NIPS2017_7017} we %generate preferences using 
use the true reward $r$ to simulate human preferences. %This %reduction generates 
%  leads to a PAF with half as many attacks as the AAF from Step 2
%  .
%We take each attack in the PAF as a preference over trajectories, i.e., for all $\tau_1$ attacks $\tau_2$ we take $\tau_1$ \textit{preferred} to $\tau_2$ and we train the reward model to predict these preferences. 

\emph{Human Preferences.} To collect HF we present a short video of the maze with two agents rolling out their trajectories. %, one in red and one in blue. 
We then ask the human, $\mathcal{H}$, to select a preference over the trajectories. %, otherwise $\mathcal{H}$ may indicate no preference.
%From the $100$ trajectories, we 
We take a random sample of attacks between the $100$ trajectories and collect 200 preferences% (we compare $\hat{r}_h$ trained on $100$ and $200$ preferences)
.

\emph{Step 4%: Generalising the Preferences
.}  We generalise both the synthetic and human preferences over trajectories to preferences over preferred extensions, as described in Section~\ref{sec:theory_steps}. To order the preferred extensions of the AAF in order to generalise the (synthetic or human) preferences, we perform a binary insertion sort, which %is optimal ( $O(n log n)$ ) w.r.t. the number of comparisons to be made, and therefore  
makes the minimum number of %queries to $\mathcal{H}$.
comparisons.
\emph{Generalising Synthetic Preferences:} We order the preferred extensions by the total return of the trajectories in them, that is, for preferred extensions $p_i$ and $p_j$:    $p_i \gg p_j \text{ iff } \sum_{\tau \in p_i} \hat{R}(\tau) > \sum_{\tau \in p_j} \hat{R}(\tau).$ 
% \begin{align}
%     p_1 \succ p_2 \text{ iff } \sum_{\tau \in p_1} R(\tau) > \sum_{\tau \in p_2} R(\tau).
% \end{align}
% Then we label each trajectory with it’s ``best” preferred extension and we remove all attacks which are from arguments in worse preferred extensions against arguments in better preferred extensions. 
Then we take $\tau_1 \succ \tau_2$ as described in Section~\ref{sec:theory_steps}. %iff $\exists p_1$ with $\tau_1 \in p_1$ s.t. $\forall p_2$ with $\tau_2 \in p_2: p_1\succ p_2$. %\nb{I do not understand the following sentence: it seems to indicate that you obtain preferred extensions first - of what? - and then obtain the preferences to obtain the PAF. May be you compute the preferred extensions of the AAF first? given that you have two arg frams around, you cannot leave these considerations implicit... }
This gives us our PAF from an order $\gg$ over preferred extensions.
\emph{Generalising Human Preferences:} % To make this experiment fair, we use the same human preferences as when training $r_h$. 
To determine the preference  between preferred extensions $p_i$ and $p_j$, we count the number of %labelled
preferences between the trajectories %of
in $p_i$ and $p_j$, as follows: %i.e. 
%attacks_{p_1,p_2} 
let $count_{p_{i},p_{j}}= \sum_{\tau_1 \in p_{i}, \tau_2 \in p_{j}} \mathds{1}(\tau_1 \succ \tau_2)$%, for $i,j=1,2, i \neq j$
; where $\mathds{1}$ is the indicator function.
% \begin{align}
%     attacks_{p_1,p_2} = \sum_{\tau_1 \in p_1, \tau_2 \in p_2} Bool(\tau_1 \succ \tau_2)
% \end{align}
Then $p_i %\succ
\gg p_j$ iff $%attacks
count_{p_i,p_j} > %attacks
count_{p_j,p_i}$. %This is a simple method to order the preferred extensions using the collected HF. 
We generate the PAF following the approach in Section~\ref{sec:theory_steps}.  %We then reduce the attacks as above. 

\emph{Step 5%: Training the Reward model $\hat{r}$
.} We %now train the reward model using 
follow the process described in Section \ref{sec:5}. 
%The hyper-parameters are outlined in Appendix \ref{sec:detailed_res}.

% Recall that the reward model takes as input a state and action and predicts the immediate reward. We generate a preference prediction from rewards with

% \begin{align}
% P(\tau_1 \succ \tau_2) = \frac{\exp(R(\tau_1))}{\exp(R(\tau_1)) + \exp(R(\tau_2))}
% \end{align}  

% and the loss to the reward model is then $-log(P(\tau_1 \succ \tau_2))$ . 

\emph{Step 6%: Training the Policy $\pi$
.} For each reward %function 
model resulting from Step 5 %(depending on whether we use synthetic or human preferences)}
we use the RL algorithm as indicated in Section~\ref{sec:theory_steps} %and hyper-parameters described in appendix \ref{sec:hyper} 
to train a policy. %In section \ref{sec:map2_results} we compare the policies learned. 

\emph{Iterative Reward Modelling.}
We cycle through steps 1-6 to iteratively train a reward model and policy from HF, with and without generalisation. %\del{full details in Appendix \ref{sec:iter})}.
At the first iteration we generate random trajectories,  learning a reward model and policy as in the non-iterative version; then, at every following iteration,  we start with the trajectories generated at the previous. We increase trajectory length  with iterations, with each trajectory starting from the same initial state. 
 %Specifically, we include an initial stage where only the reward model is updated (not the policy) to prevent a bad policy being learned from the uninitialised (random) reward. %We collect $10$ trajectories at a time, starting with an episode length of $20$ and increasing this by $5$ every step. 
 %Each trajectory starts from the same initial state and we gradually increase the trajectory length each iteration. We query the human for preferences between attacking trajectories collected at each iteration. %(we do not query attacks between trajectories gathered at different iterations). 
When we train the iterative reward model \emph{without generalisation}, we query a random sample of ten %attacks 
pairs of trajectories at each stage, or all of the %attacks 
pairs if there are fewer than ten. When we train the iterative reward model \emph{with generalisation}, we make the minimum %\nb{what does this mean?}
number of queries to order the preferred extensions. %We show the human two trajectories at a time and ask for a preference, otherwise the human can label no preference (in which case the attack is discarded). Then these preferences are used to train the reward model. %At each step we train the reward model for $10$ epochs with batch size equal to the number of labelled preferences. 
In each case we %continue this process 
iterate for   15 minutes.  %This process continues for $15$ minutes collecting a total of $131$ preferences.

%%%%%%%%%%%5

\begin{figure}[t]
     \centering
     \begin{subfigure}[b]{0.23\textwidth}
         \centering
         \includegraphics[width=\textwidth]{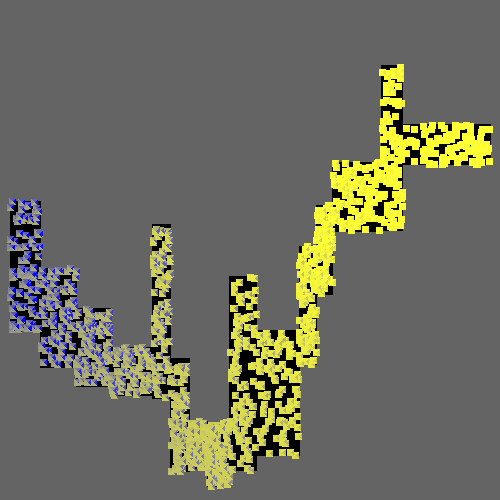}
         \caption{Standard RLHF ($r_{h, 100}$).}
         \label{fig:rh100_total}
     \end{subfigure}
     \hfill
     \begin{subfigure}[b]{0.23\textwidth}
         \centering
         \includegraphics[width=\textwidth]{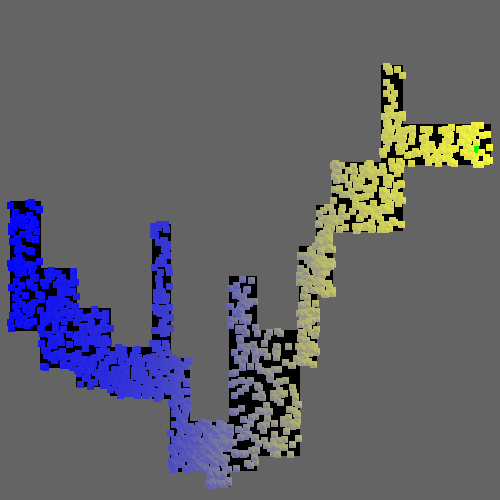}
         \caption{ARL ($r_{h, g, 100})$.}
         \label{fig:rhg_total}
     \end{subfigure}
        \caption{Reward heatmaps normalised over whole state space (yellow is high and blue is low reward).}
        \label{fig:heat1}
\end{figure}
% \vspace{-10mm}
\begin{figure}[t]
     \centering
     \begin{subfigure}[b]{0.23\textwidth}
         \centering
         \includegraphics[width=\textwidth]{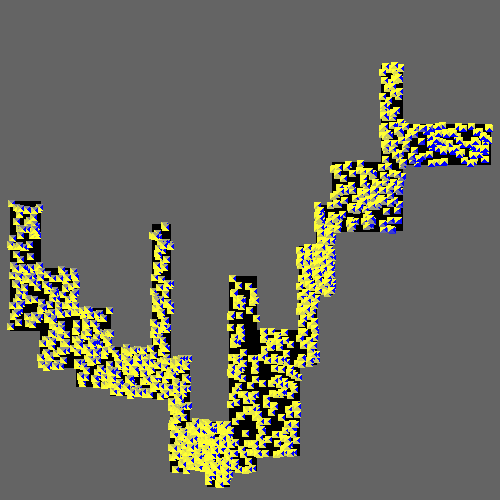}
         \caption{Standard RLHF ($r_{h, 100}$).}
         \label{fig:rh100_action}
     \end{subfigure}
     \hfill
     \begin{subfigure}[b]{0.23\textwidth}
         \centering
         \includegraphics[width=\textwidth]{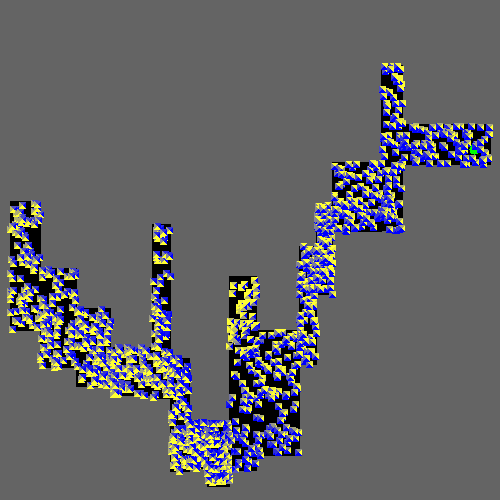}
         \caption{ARL ($r_{h, g, 100})$.}
         \label{fig:rhg_action}
     \end{subfigure}
        \caption{Reward heatmaps showing best action at each point.}
        \label{fig:heat2}
\end{figure}

% \vspace{10mm}
\section{Results } \label{sec:results}

\begin{table}[t!]
\caption{\emph{Performance of Reward models.} 
Our reward models highlighted in blue.
$r$ is the true reward and $\hat{r}$ is the learnt reward, with
$s$ = synthetic preferences, $h$ = human preference, $g$ = generalised, $i$=iterated. 
We present the number of preferences used to train the model (and, if the preferences were generalised, the number of human labels used), the mean preference prediction accuracy (MPPA) on a test set, and the distance to the goal achieved by each policy.
 A lower distance to goal does not necessarily correspond to a better policy, since the agent may need to travel further away from the goal to navigate deeper into the maze, therefore we highlight ``good" runs (meaning that the agent moves meaningfully through the maze)
in green and ``bad" runs (in which the agent does not move meaningfully through the maze) in red.
% \nb{I do not understand how 15 mins for the preferences and the minutes for distance to the goal interplay}
}
\label{tab:map2}
\centering
\resizebox{0.49\textwidth}{!}{
\begin{tabular}{|c|c|c|c|c|c|}
\hline
& & & \multicolumn{3}{c|}{Distance to goal (normalised to $[0,1]$) achieved by policy trained for 5/10/15 minutes }   \\
\hline
Reward & $\#$ Preferences & MPPA (Test) & 5 & 10 & 15   \\
\hline

$r$ & n/a & n/a & 0.725  $\pm$ 0.013 &  0.654 $\pm$ 0.0058 & 0.646 $\pm$ 0 \\

$\hat{r}_{s}$ & 4115  &  0.895 & \color{red} 0.841 $\pm$ 0 & \color{red} 0.841 $\pm$ 0  & \color{red} 0.841 $\pm$ 0 \\

\cellcolor{blue!25} $\hat{r}_{s,g}$ & 4115 & 0.954 & \color{green} 0.766 $\pm$ 0.0613 & \color{green}  0.675 $\pm$ 0.0432 & \color{green}  0.7459 $\pm$ 0.0625 \\

$\hat{r}_{h}$ & 100 & 0.798 & \color{red} 0.969 $\pm$ 0  & \color{red} 0.969 $\pm$ 0 &  \color{red} 0.969 $\pm$ 0 \\

\cellcolor{blue!25} $\hat{r}_{hg}$ & 4115 (from 100) & 0.847 &\color{green}  0.694 $\pm$ 0.116  &  \color{green} 0.632 $\pm$ 0.0234  &  \color{green} 0.689 $\pm$ 0.0340  \\

$\hat{r}_{h}$ & 200 & 0.896 & \color{green} 0.937 $\pm$ 0.0227  & \color{green} 0.8996 $\pm$ 0.008095 & \color{green} 0.9259 $\pm$ 0.0271 \\

\cellcolor{blue!25}  $\hat{r}_{hg}$ & 4115 (from 200) &  0.893 & \color{green} 0.862 $\pm$ 0.0293 & \color{green} 0.495 $\pm$ 0.0868 & \color{green} 0.499 $\pm$ 0.0267  \\

%$\hat{r}_{h}$ & 300 & 0.945 &  0.969 $\pm$ 0  & 0.969 $\pm$ 0 &  0.969 $\pm$ 0 \\

%$\hat{r}_{hg}$ & 4115 (from 300) & 0.935 &   & &   \\

$\hat{r}_{h, i}$ & 131 (15 mins) & n/a & 0.738 $\pm$ 0  & 0.738 $\pm$ 0  & 0.762 $\pm$ 0.0503  \\

\cellcolor{blue!25} $\hat{r}_{h, i, g}$ &  773 (15 mins) & n/a & \color{red} 0.969 $\pm$ 0  & \color{red} 0.969 $\pm$ 0  & \color{red} 0.969 $\pm$ 0.  \\
\hline
\end{tabular}}
\vspace{-10mm}
\end{table} 
% \vspace{-10mm}
% \begin{figure}[t!] 
% \centering
% \includegraphics[width=0.5\textwidth]{pics/MPPA_smoothed.png}
% \caption{The (smoothed) mean preference prediction accuracy (MPPA) for each trained reward model: $\hat{r}_{s}$ is trained from synthetic preferences generated by the true reward function $r$; $\hat{r}_{s, g}$ is trained from synthetic preferences generated by ordering the preferred extensions using $r$; $\hat{r}_{h}$ is trained from (either $100$ or $200$) human-labelled preferences; $\hat{r}_{h, g}$ is trained from preferences generated by ordering the preferred extensions using HF.}
% \label{fig:mppas}
% \end{figure}

We compare the performance of the reward models described in the previous section
%\del{by examining reward heatmaps over the state space, the accuracy with which they predict preferences, and the policies which they generate.} %These experiments are performed on the map shown in figure \ref{fig:ded}. 
in Table \ref{tab:map2}. %\del{ presents the results for each reward model}. 
In each case we train the policy three times %, since the RL policy training is inherently stochastic, hence we 
to collect mean (and standard deviation) distance to goal which the resultant greedy Q-network achieves.
% We first note that the distance to the goal is the primary value with which the designed reward, $r$, is parameterised; this can be considered the main weakness of the designed reward, compared to the reward models trained on HF which provide reward for actions which correspond to genuinely moving deeper into the maze. However, w
% \nb{what about Fig.~\ref{fig:mppas}? need to point to it and somewhat, even minimally, discuss here}
% In Fig.~\ref{fig:mppas} we present the accuracy with which the reward models predict preferences during training, it can be seen that using our method of generalisation increases the initial rate of learning on the experiments with genuine human feedback. Although, overall the instability of training the deep reward models makes it difficult to draw concrete conclusions. 
We also %assess the accuracy with which different reward models predict preferences and the policies which they generate by 
examine reward heatmaps over the state space (see Fig.~\ref{fig:heat1} and Fig.~\ref{fig:heat2}) in order to visualise what the reward models have actually learned.

We note that the reward model trained on $100$ human preferences without generalisation, $r_{h, 100}$, over-fits the training data and gives high reward for moving down near the initial state (and we can observe from Fig. \ref{fig:rh100_total} that the reward is not well distributed throughout the state-space%, with high reward given for moving down in the initial region
). Hence, the policy trained on $r_{h, 100}$ learns to just move downwards. When we generalise these same $100$ preferences %by ordering the preferred extensions, 
we observe much better performance (MPPA $0.847$ compared to $0.798$). We see in Fig. \ref{fig:rhg_total} that the reward is more sensibly distributed around the state space and we see by comparing Fig. \ref{fig:rh100_action} to Fig. \ref{fig:rhg_action} that generalising preferences improves the reward model's predicted best action in each state. As expected, using $200$ preferences instead of $100$ improves performance, and generalisation again improves the policy (the MPPA is around $0.89$ for both $r_{h, 200}$ and $r_{h, g, 200}$). Similarly, generalising synthetic preferences improves both MPPA and the policy.
% From this, and the improved MPPA and policy resulting from generalising the synthetic preferences, we may tentatively conclude that generalising the preferences using the preferred semantics leads to genuine performance improvements, especially when there is only a small amount of preference-data available with which to train the reward model. 
Surprisingly, %the experiments with the full iterative process are somewhat contradictory to this conclusion, as 
in the iterative learning experiments, generalisation decreases performance. However, these experiments do not use fixed trajectories or HF (since the point of the iterative process is to generate increasingly better trajectories based on the current reward).  %The discrepancy here could be due to a number of factors, such as the stochasticity in the exploration, or the particular hyper-parameters used. 

% \subsubsection{Hyper-Parameters}

% In general, finding a combination of hyper-parameters across all different methods is difficult. That is, we need a good combination of hyper-parameters across the policy learning ($\epsilon$, trajectory length, etc), reward predicting (learning rate, number of epochs, etc), and generalising algorithms (attack similarity-threshold, number of queries etc). In particular, there is no such thing as a good hyper-parameter value, only good combinations of hyper-parameters, and so, when comparing reward functions by the policy they produce one reward may be better suited to the particular RL algorithm and hyper-parameters used. It is therefore difficult to conduct perfect experiments which truly fairly compare reward functions in this way. 

%  We also saw that it is difficult to compare reward functions fairly, for instance, because the performance is highly sensitive to the combination of \\ hyper-parameters across methods, and because the absolute value of rewards is not relevant. Furthermore, in some circumstances, the resultant policy may actually perform better (by some metric of the policy) precisely because the reward-model does a worse job of capturing the revealed preferences. 


\section{Conclusion}

%Our contribution is 
\emph{Summary.} We presented a novel neuro-symbolic framework, \emph{argumentative reward learning} (ARL). ARL incorporates argumentative reasoning into the reinforcement learning from human feedback loop, in order to generalise human preferences. ARL improves past work, increasing the accuracy with which the reward model predicts preferences (especially when there is only limited human feedback available) in addition to improving the learned policy. 

\emph{Future work.} %We would be excited to see future work developing further methods for incorporating symbolic methods into RLHF. In particular, o
Other methods for reasoning about human preferences could be explored, and more principled methods for active learning might be developed (e.g. in which a dialogue between the system and human in the loop can be used to extract the most useful information about the human's preferences). %Alternatively, techniques for formal verification could be used to verify that learned rewards genuinely capture human preferences. 

% \nb{add explicitly which future work about the proposed approach would be meaningful: dropping some of the restrictions? considering other environments? other RL settings? }

% \nb{the references need fixing: uniformity and drop some unnecessary info (e.g. month of publication, location of conferences)}

\bibliographystyle{icml2022}
\bibliography{arg_reward_learning}

\end{document}